%% file: neurips_2023.tex
\documentclass{article}

\PassOptionsToPackage{numbers, compress}{natbib}

\usepackage[preprint]{neurips_2023}

\usepackage[utf8]{inputenc} 
\usepackage[T1]{fontenc}    
\usepackage{hyperref}       
\usepackage{url}            
\usepackage{booktabs}       
\usepackage{amsfonts}       
\usepackage{nicefrac}       
\usepackage{microtype}      
\usepackage{xcolor}         
\usepackage{graphicx}
\usepackage{multicol}
\usepackage{amsmath}
\usepackage{multirow}

\title{Mipha: A Comprehensive Overhaul of Multimodal Assistant with Small Language Models}

\author{%
Minjie Zhu$^{1,2}\thanks{Equal contribution. $\dagger$ Project lead: yichen\_zhu@foxmail.com}$ \quad Yichen Zhu$^{1*\dagger}$ \quad Xin Liu$^2$ \quad Ning Liu$^2$ \quad Zhiyuan Xu$^2$ \\
\textbf{Chaomin Shen}$^2$ \quad \textbf{Yaxin Peng}$^3$ \quad \textbf{Zhicai Ou}$^1$ \quad \textbf{Feifei Feng}$^1$ \quad \textbf{Jian Tang}$^1$\\
$^1$Midea Group \quad $^2$East China Normal University \quad $^3$Shanghai University\\
}

\begin{document}

\maketitle

\input{abstract}
\input{intro}
\input{related_work}
\input{method}

\input{experiment}
\input{conclusion}

\bibliographystyle{plainnat}
\bibliography{egbib.bib}
\input{appendix/appendix}

\end{document}

%% file: abstract.tex
\begin{abstract}
   Multimodal Large Language Models (MLLMs) have showcased impressive skills in tasks related to visual understanding and reasoning. Yet, their widespread application faces obstacles due to the high computational demands during both the training and inference phases, restricting their use to a limited audience within the research and user communities. In this paper, we investigate the design aspects of Multimodal Small Language Models (MSLMs) and propose an efficient multimodal assistant named Mipha, which is designed to create synergy among various aspects: visual representation, language models, and optimization strategies. We show that without increasing the volume of training data, our Mipha-3B outperforms the state-of-the-art large MLLMs, especially LLaVA-1.5-13B, on multiple benchmarks. Through detailed discussion, we provide insights and guidelines for developing strong MSLMs that rival the capabilities of MLLMs. Our code is available at \href{https://github.com/zhuyiche/llava-phi}{https://github.com/zhuyiche/llava-phi.} 
\end{abstract}

%% file: intro.tex
\section{Introduction}
\label{sec:intro}

Recent advancements in Multimodal Large Language Models (MLLMs) have demonstrated exceptional visual understanding and reasoning performances across a range of tasks such as visual question answering~\cite{gemini,flamingo,cogvlm} and referring comprehension~\cite{gpt4}. Benefiting from the scaling law of Transformer architecture~\cite{vaswani2017attention} and web-scale training data sources, these models have become foundational in the field of artificial intelligence, with their parameters increasing from billions to trillions~\cite{gpt4, llama2, chowdhery2023palm}. However, the deployment of these models is often hindered by their substantial computational costs and memory consumption in both the training and inference phases, which limits their popularization across the broader research and user communities. 

How can we boost the inference speed of MLLMs?  Delving into the computational costs of MLLMs, it becomes evident that the Large Language Models (LLMs), primarily tasked with transforming visual-text pairs into textual outputs, account for a substantial computational load. In MLLMs, the visual encoder, such as CLIP-ViT-L, typically only comprises around 0.4 billion parameters, while the language model within MLLMs can escalate to between 7 and 65 billion parameters.

Intuitively, reducing the computational demands of the language model could lead to a significant decrease in overall inference costs. The rapid advancements by the open-source community in developing Small Language Models (SLMs), with parameter sizes ranging from 1 billion to 3 billion (i.e., Phi~\cite{phi-1,phi-1.5,Phi-2}), enable us to construct vision-language models more compact than traditional MLLMs. 
However, as the parameter count of these smaller language models is reduced, their capabilities often do not measure up to those of their larger counterparts, potentially diminishing the effectiveness of small multimodal models. This motivation drives us to explore the development of multimodal assistants using small language models that can deliver competitive performance comparable to their counterparts~\cite{instructblip,qwen,mplugowl,mplugowl2,llava1.5}, which rely on large language models.

\begin{figure}[tb]
  \centering
  \includegraphics[width=\linewidth]{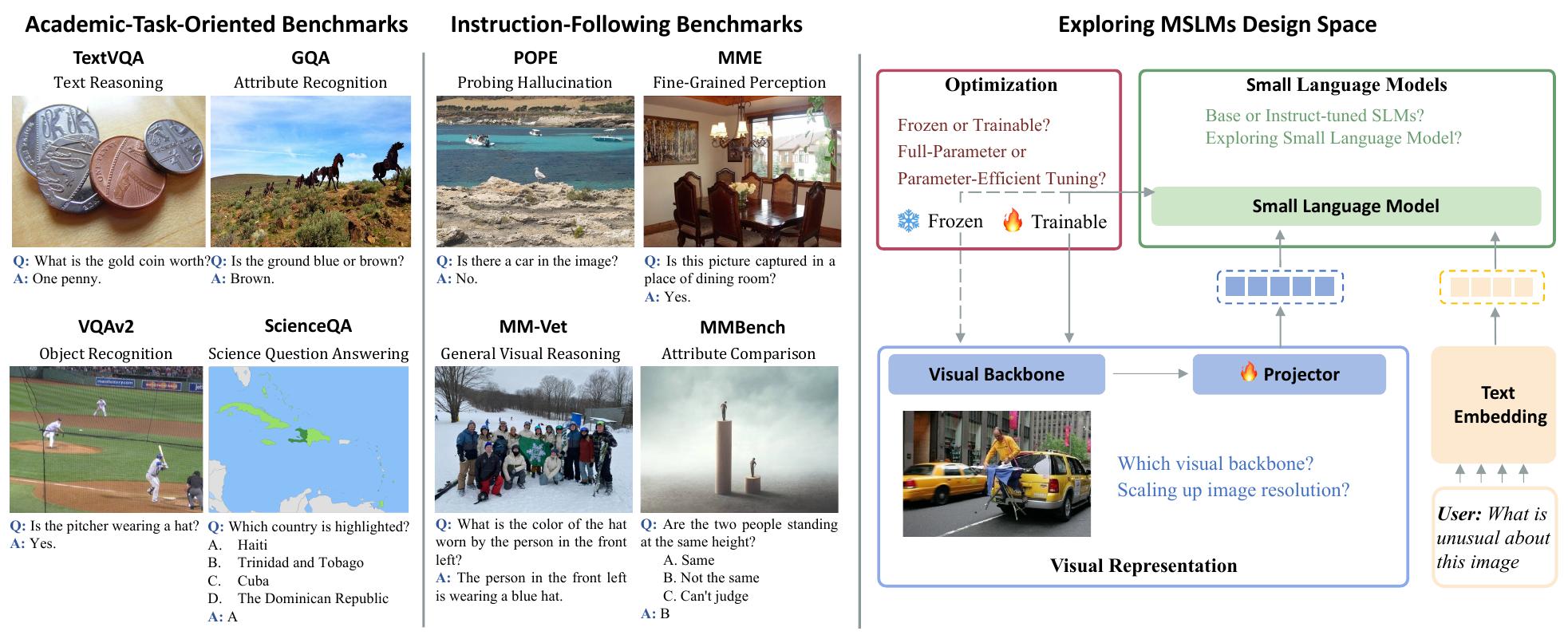}
  \caption{\textbf{Evaluation Benchmarks Overview.} We evaluate our model variants on academic-task-oriented benchmarks (\textbf{left}) as well as instruction-following benchmarks (\textbf{middle}). The answer is predicted by our proposed efficient MSLM: Mipha-3B. Additionally, we explore three key design spaces of MSLMs: 1) visual representation, 2) language model, and 3) optimization strategy (\textbf{right}).}
  \label{fig:benchmark}
\end{figure}
In this paper, we present an in-depth analysis of the impact of various design elements in MSLMs on their downstream performance. A brief summary of the design space that we explored is illustrated in Fig~\ref{fig:benchmark} (right). Our investigation dissects the existing MSLMs into three key components: the language model, the visual representation, and the optimization strategy. 
Our thorough empirical research leads us to \textbf{several findings that diverge from the conventional wisdom} established by prior studies on Multimodal Large Language Models. 
For example:
\begin{itemize}
    \item Increasing image resolution is not a silver bullet. In some benchmarks, images with a resolution of 224 pixels outperform those with 448 pixels. 
    \item Contrary to earlier findings that suggest finetuning the visual backbone negatively impacts MLLMs, we underscore the importance of simultaneously finetuning both the visual backbone and the language model for MSLMs.
    \item While the prevailing trend among MLLMs is to employ instruct-tuned language models, such as Vicuna~\cite{vicuna}, our analysis reveals that instruction tuning—be it through supervised fine-tuning or reinforcement learning from human feedback (RLHF)—is not essential.
\end{itemize}
Building on these findings, we present a new family of MSLMs named Mipha, scaled from 1.7B to 3B, which surpasses the performance of leading open-source MLLMs including LLaVA-1.5 and Qwen-VL, without the need for additional training data. Remarkably, our Mipha-3B models demonstrate superior performance across a majority of benchmarks when compared to the 7B MLLM variants, such as LLaVA-1.5, Qwen-VL, and InstructBLIP, and in some cases, even outperform 13B MLLM counterparts. Overall, we believe that our analysis provides a practical new perspective on optimizing effective MSLMs.
Overall, our contributions are the follows:
\begin{enumerate}
    \item We offer a detailed examination of the design landscape for Multimodal Small Language Models, examining visual components, language models, and optimization strategies from a comprehensive perspective. Our empirical research provides unique findings, \textbf{presenting multiple conclusions that contrast with prior studies on Multimodal Large Language Models}.
    \item By integrating our empirical discoveries, we unveil a series of Multimodal Small Language Models named Mipha. This advancement underscores the practical impact of our findings.
    \item Demonstrating the effectiveness of our approach, we show that our flagship model, Mipha-3B, outperforms existing 7B and 13B models on various benchmarks without requiring additional data for training. This marks a significant advancement for multimodal models with fewer than 3 billion parameters, establishing a new benchmark in the field.
\end{enumerate}

%% file: related_work.tex
\section{Related Work}
\textbf{Language Models.} Arming with scaled-up data and models, Large Language Models (LLMs) have demonstrated emergent capabilities like zero-shot generalization and in-context learning ability~\cite{chowdhery2023palm,flant5,gpt4,llama,llama2}. 
Despite these advances, LLMs are computationally intensive, demanding significant computational resources for training and inference. This backdrop has inspired the development of smaller language models~\cite{phi-1,phi-1.5,Phi-2,pythia,mobilellm,tinyllama,stablelm,gemma}, which, despite having fewer than 3 billion parameters, exhibit strong generalization capabilities in language understanding and question-answering tasks, rivaling their larger counterparts such as 7 billion parameter models. 
In our research, we explore the integration of these compact language models with visually-conditioned models, aiming to empower neural networks with the dual ability to comprehend and interpret both textual and visual information.
\\
\\
\noindent
\textbf{Multimodal Large Language Models.} Multimodal Large Language Models (MLLMs) connect vision and language and extend the reasoning ability of LLMs to process with multimodal input.  Numerous works have been proposed in this direction~\cite{blip-2,instructblip,minigpt4,llava,gemini,minigptv2,otter,sphinx,fuyu-8b}, which most works differ based on their adapter design, training strategy, instruction-tuning/pretraining datasets. Leveraging their complex reasoning and advanced understanding abilities, LLMs serve as the "brain" of these systems, interpreting visual elements based on text queries and generating coherent, fluent textual responses in dialogues.  Currently, the bulk of the computational demand stems from this "brain", with parameter counts ranging from 7 billion to 65 billion for publicly available models. 
This substantial requirement for computational resources presents a significant challenge for deployment on mobile or edge devices, making the inference cost-prohibitive and limiting the accessibility of MLLMs to a wider audience. This work aims to explore MSLMs that can achieve performance competing with MLLMs over 7B parameters. 
\\
\\
\noindent
\textbf{Multimodal Small Language Models.} Recently, a select number of studies~\cite{imp,bunny,mobilevlmv2,llavaphi,moe-llava,vary-toy,zhang2022opt} have delved into the exploration of Multimodal Small Language Models (MSLMs) from diverse angles. Several works, i.e, LLaVA-Phi~\cite{llavaphi}, utilize pretrained small language models as the core of multimodal models to reduce computational demands. The MobileVLM~\cite{chu2023mobilevlm,mobilevlmv2} series focuses on the innovative design of projectors to enhance the performance of MSLMs. MiniCPM-V~\cite{minicpm2024} seeks to refine the training approach, and Bunny~\cite{bunny} examines the effects of augmenting the volume of training data. Our paper put a thorough investigation into the design space for training an effective MSLM. 

%% file: method.tex
\section{Preliminaries}
In this section, we provide an overview of model architecture and evaluation benchmarks.
\subsection{Overview of Model Architecture} 
The architecture of mainstream MLLMs \cite{llava1.5,minigpt4,mplugowl2} typically comprises three primary components: 1) a visual representation backbone $V_{\psi}$, 2) a vision-language projector $F_{\omega}$ aimed at aligning visual and textual domains, and 3) a language model $LM_{\theta}$.
\\
\\
\noindent
\textbf{Visual Representation Backbone.} Given an input image $X_{img}$, we consider a visual representation backbone to output the visual features $V_{img}\in \mathbb{R}^{L \times h_{v}}$, where $V_{img}=V_{\psi}(X_{img})$. The $L$ denotes the number of patches. All of our experiments consider only the grid features before and after the last transformer layer.
\\
\\
\noindent
\textbf{Vision-Language Projector.} Then, a vision-language projector is employed to convert the visual features $V_{img}$ to the text embedding space $T_{img} \in \mathbb{R}^{L \times h_{t}}$, where $T_{img}=F_{\omega}(V_{img})$. Following LLaVA-1.5~\cite{llava1.5}, we employ a two-layer MLP as our projector across all of our model variants, which can effectively align visual and textual features.
\\
\\
\noindent
\textbf{Language Model (LM).} Finally, based on $T_{img}$ and the question embedding $E_{t}$, a language model is used to give the answer $A$, where $A=LM_{\theta}(text{cat}(T_{img}, E_{t}))$. While previous studies mainly focus on language models with more than 7 billion parameters, such as Vicuna or LLaMA, our work shifted attention to multimodal models that are equipped with SLMs. 
\begin{figure}[tb]
  \centering
  \includegraphics[width=0.95\linewidth]{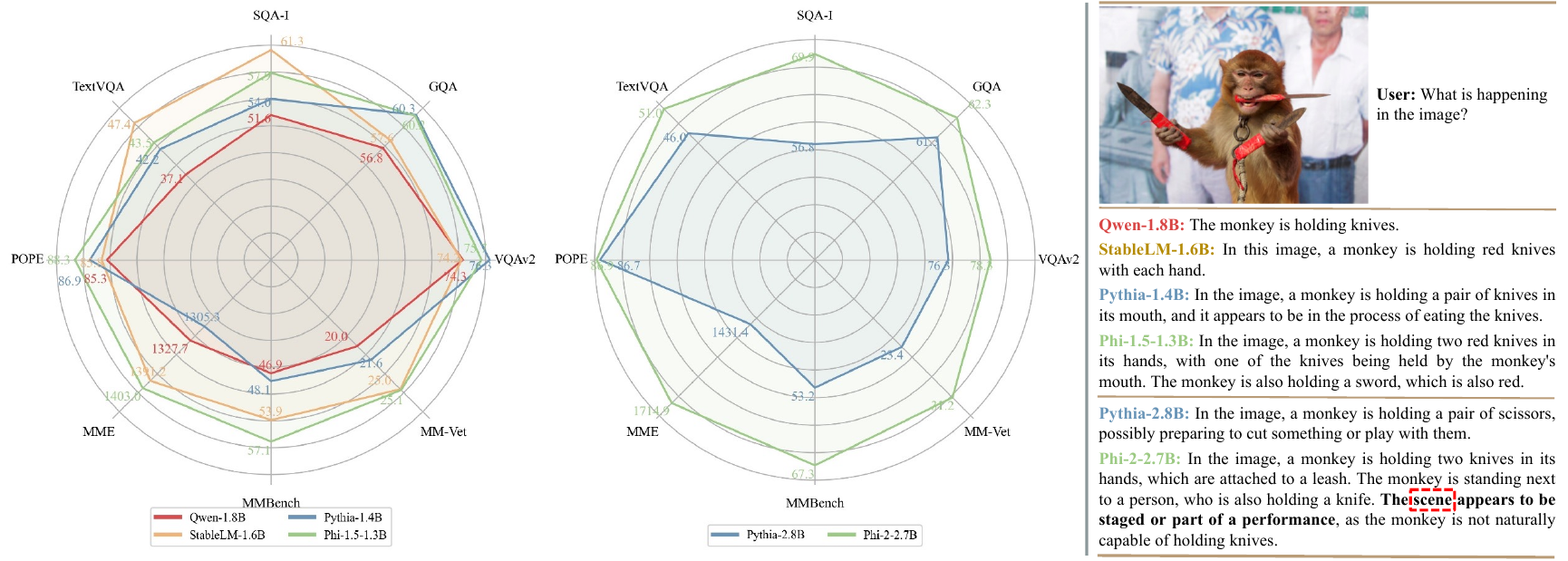}
  \caption{\textbf{Selection of Small Language Models.} We evaluate four open-sourced SLMs on 8 benchmarks, and Phi-family showcases the best performance (\textbf{left $\&$ middle}). We discovered that MSLMs equipped with Phi-2-2.7B is able to identify that the monkey in the image is performing, a subtlety that other models fail to recognize (\textbf{right}).}
  \label{fig:slm}
\end{figure}
\input{slm_table}

\subsection{Evaluation Benchmarks.}
We conduct empirical studies on a collection of both academic-task-oriented benchmarks and recent benchmarks specifically proposed for instruction-following MLLMs, totaling 8 benchmarks. For academic-task-oriented benchmarks, VQA-v2~\cite{vqav2} and GQA~\cite{gqa} evaluate the model’s visual perception capabilities on open-ended short answers. ScienceQA~\cite{scienceqa} is used to evaluate the zero-shot generalization on scientific question answering. TextVQA~\cite{textvqa} contains text-rich visual question answering. We also employ recent benchmarks proposed for instruction-following MLLMs. POPE~\cite{pope} evaluates the model’s degree of hallucination on three sampled subsets of COCO~\cite{coco}. MME~\cite{mme} Benchmark evaluates MLLMs’ perception and cognition capabilities, and MMBench~\cite{mmbench} evaluates the model’s answer robustness with all-around shuffling on multiple-choice answers.
MM-Vet~\cite{mmvet} evaluates the model’s capabilities in engaging in visual conversations on a diverse range of tasks. Some examples are illustrated in Fig~\ref{fig:benchmark}.

\section{Exploring Crucial Aspects on Efficient MSLMs}
There are a few prior studies that have explored different aspects of enhancing MLLMs~\cite{karamcheti2024prismatic,scalingmllm}. However, these works primarily focus on multimodal assistants powered by large language models. Intuitively, the model insights obtained from multimodal large language model should also apply to multimodal smaller language models. Nevertheless, in the following section, the contrary: techniques effective for large-scale models do not always seamlessly transfer to small-scale models. We also demonstrate several new observations that can improve the performance of MSLMs. We highlight that all our experiments in this section utilize the same amount of training data, and unless otherwise stated, all training components are maintained consistently for a fair comparison.

\subsection{What makes Small Language Model good for Multimodal Assistant?}
In this section, we delve into the methodology for selecting and training an effective Small Language Model (SLM) that enhances the performance of Multimodal Small Language Models (MSLMs). Our discussion is twofold. Initially, we evaluate various SLMs to determine which model exhibits optimal performance upon integration with a visual backbone. Subsequently, we investigate the MSLMs' performance with base or instruct-tuned language models.

\subsubsection{Which Small Language Model should we adopt?} As the "brain" of multimodal models, the selection of pretrained small language models is crucial for crafting high-performing Multimodal Small Language Models (MSLMs).
We evaluate a spectrum of recently introduced Small Language Models (SLMs) with sizes ranging from 1.3B to 2.8B parameters. This includes models such as Phi-2 (2.7B)~\cite{Phi-2}, Phi-1.5 (1.3B)~\cite{phi-1.5}, Qwen-1.5 (1.8B)~\cite{qwen}, Pythia (available in both 1.4B and 2.8B sizes)~\cite{pythia}, and StableLM (1.6B)~\cite{stablelm}. For analytical clarity, we classify these SLMs into two categories: those models possess over 2.7B parameters, and the remainder, feature less than 2B parameters.

\begin{figure}[tb]
  \centering
  \includegraphics[width=\linewidth]{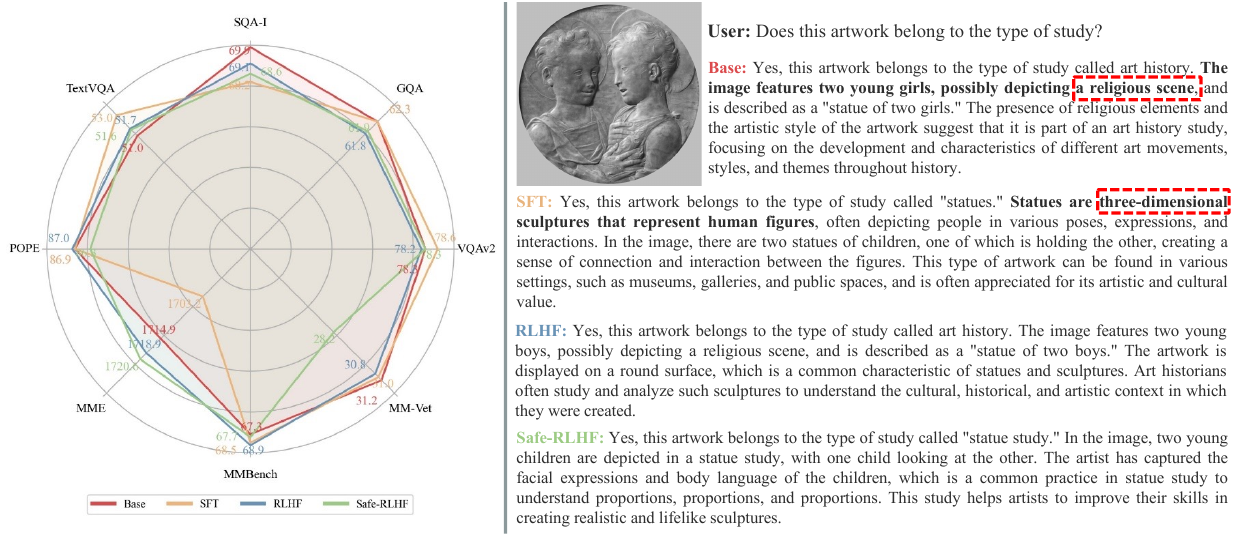}
  \caption{\textbf{Base vs. Instruct-tuned LMs.} For MLLMs, we explore the differences using base LM and instruct-tuned LM. While the quantitative performance metrics appear similar, the qualitative results reveal differences (\textbf{left}). For example, when comparing the responses generated by MLLMs equipped with either a Base or SFT LM to those from models finetuned with RLHF or Safe-RLHF, the latter is observed to be more verbalized (\textbf{right}).}
  \label{fig:vanilla_sft_rlhf_safe_rlhf}
\end{figure}
As illustrated in Figure~\ref{fig:slm} (left), we assess the performance of MSLMs that incorporate language models with fewer than 2 billion parameters. Phi-1.5-1.3B, despite having fewer parameters compared to its counterparts, still manages to display competitive or even superior performance. To delve into the reasons behind Phi-1.5's outperformance, we benchmark these SLMs against four tests traditionally used for LLMs, as depicted in Table~\ref{tbl:slm_table}. These benchmarks include the MMLU for language understanding, BBH for instruction following, GSM8K for mathematics, and HumanEval for coding. Our observations reveal that Phi-1.5-1.3B remains competitive or outperforms models with a larger parameter count, such as Pythia-1.4B, StableLM-1.6B, and Qwen-1.8B. This suggests that Phi-1.5-1.3B makes efficient use of its parameters, particularly in the areas of reasoning and language understanding that these benchmarks measure, enhancing its transferability to multimodal contexts. Our findings indicate that a strong small language model can lead to better multimodal performance. 

Moreover, when comparing Phi-2-2.7B with Pythia-2.8B, two relatively larger SLMs, the superiority of Phi-2 is evident. Aside from POPE, where models with Pythia show slightly fewer object hallucinations, Phi-2 models achieve markedly better outcomes on other metrics. This aligns with expectations, as Phi-2 was released after Pythia and was pretrained on a considerably larger dataset—1.4 trillion tokens versus 300 billion tokens.

Additionally, the qualitative results corroborate our findings. MSLMs based on the Phi models provide detailed descriptions derived from the images. Notably, the MSLM equipped with Phi-2-2.7B goes a step further by recognizing that the monkey is performing, which showcases a robust capacity for scene reasoning. Additional examples and detailed analyses are provided in the appendix.
\\
\\
\noindent
\textbf{Base vs. Instruct-tuned LMs.} The most state-of-the-art MLLMs~\cite{llava1.5,instructblip} leverage instruct-tuned LM, such as Vicuna, as the default backbone. However, employing instruction tuning on LMs means extra training time, and it also brings bias and regressions in performance~\cite{rlhf}. As such, we investigate whether instruction tuning is necessary for MSLMs. We consider the following four settings:
\begin{itemize}
    \item \textbf{Base.} The base language model (LM) inherits the weights from pretrained models as is. Typically, this base LM exhibits limited proficiency in following instructions, due to its training on a general corpus without specific emphasis on instruction adherence.
    
    \item \textbf{Supervised Fine-tuning (SFT).} Supervised Fine-Tuning (SFT) refines a pretrained model's parameters using a targeted, often smaller, dataset with labeled examples to enhance its performance for a specific task. This technique builds upon the broad capabilities acquired during the model's initial training phase, sharpening them to boost accuracy, relevance, or suitability for a designated domain or application. Within the realm of Multimodal Large Language Models (MLLMs), employing a language model enhanced through SFT, such as Vicuna, serves as a prevalent strategy.

    \item \textbf{Reinforcement Learning from Human Feedback (RLHF).} RLHF~\cite{rlhf} finetunes LMs by integrating preferences or feedback directly from humans. This method refines the capabilities of base LMs to function effectively as dialogue agents. This approach not only enhances the responsiveness and relevance of the LMs to user queries but also aligns their outputs more closely with human values and expectations, thereby broadening their utility as interactive tools.
    
    \item \textbf{Safe Reinforcement Learning from Human Feedback (Safe-RLHF).} Safety alignment is a critical step to mitigate the risk of LMs generating undesirable outputs in response to certain prompts~\cite{liu2023query}. In our approach, we employ Safe-RLHF~\cite{safe-rlhf}, a variant of RLHF, which incorporates safety preferences throughout the RLHF process. This method is designed to ensure that the model's responses adhere to safety guidelines, effectively preventing inappropriate outputs.
    
\end{itemize}

The experimental findings are presented in Fig~\ref{fig:vanilla_sft_rlhf_safe_rlhf} (left), revealing that there is no universal solution to enhancing model generalization.
The comparative analysis across all 8 benchmarks shows minimal variance among the four configurations tested. 
Notably, the MSLM leveraging the base Phi-2 model excels in the SQA-I and MM-Vet benchmarks. 
This superiority can likely be attributed to Phi-2's extensive pretraining on large code datasets, which enhances its ability to generalize in science question-answering contexts and generate open-ended responses. 
While the performance differences among instruct-tuned models are modest, RLHF emerges as the most versatile performer across the eight benchmarks.
Despite the close numerical results, qualitative analysis reveals a distinct divergence between instruct-tuned and base models. 
As depicted in Fig~\ref{fig:vanilla_sft_rlhf_safe_rlhf} (right), when tasked with describing an artwork image, the responses from models with base LM and SFT LM adopt a formal tone, employing terms like "religious elements" and "three-dimensional sculptures", reminiscent of textbook descriptions.
Conversely, models finetuned with RLHF and Safe-RLHF LMs offer descriptions in a more human-like, conversational manner, akin to a museum commentary.
\\
\\
\noindent
\textit{Takeaway.} \textbf{The benefit of employing an instruct-tuned LM for MSLMs over a base LM is not necessarily clear-cut}, irrespective of the instruction tuning approach used, whether it be supervised fine-tuning, reinforcement learning from human feedback (RLHF), or safety-aware RLHF. 
This finding contrasts with prior research, which indicated that using an instruct-tune language model like Vicuna offers advantages over a base language model like LLaMA.
However, \textbf{qualitative differences become apparent}: MSLMs developed using RLHF-based instruction tuning methods tend to be more verbose and produce responses that are more akin to human-like answers.
\begin{figure}[t]
  \centering
  \includegraphics[width=0.95\linewidth]{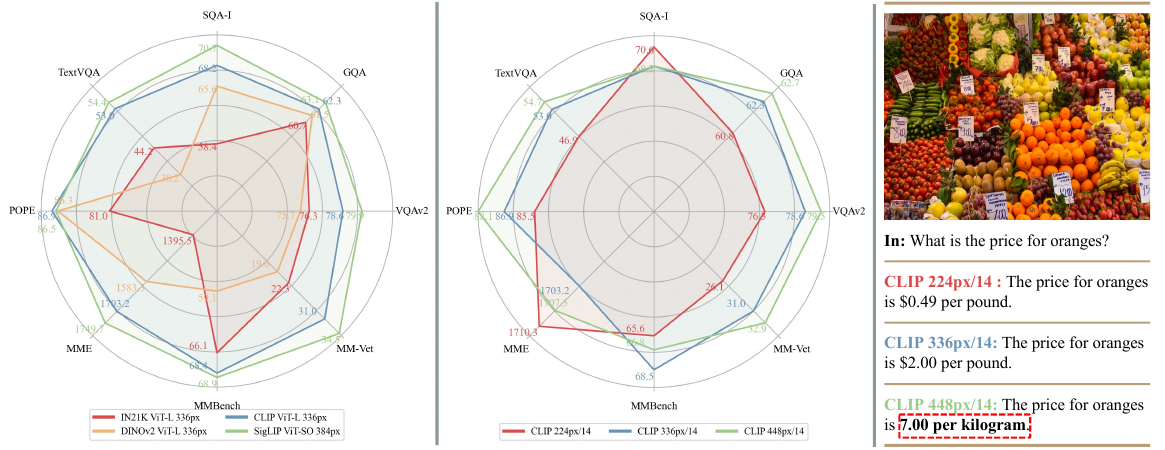}
  \caption{\textbf{Choosing a Pretrained Vision Representation \& Scaling Image Resolution.} We evaluate various visual backbones, such as CLIP, SigLIP, DINOv2, and ViT-IN21K \textbf{(left)}. We analyze model performance in relation to increasing image resolution \textbf{(middle)}. We provide qualitative examples from multimodal models employing visual backbones at different image resolutions \textbf{(right)}.}
  \label{fig:clip_siglippx}
\end{figure}

\subsection{Visual Representations in MSLM}
\textbf{Pretrained Visual Representation.} In multimodal models, a common approach is to employ a pretrained visual backbone that converts image features into latent embeddings. The quality of these pretrained embeddings is pivotal to the success of multimodal learning tasks. We evaluated the efficacy of four different pretrained visual backbones: 1) CLIP~\cite{clip}, trained on a large proprietary dataset by OpenAI, 2) SigLIP~\cite{siglip}, pretrained on the WebLI dataset, 3) ViT with IN21K, trained on the ImageNet21K~\cite{imagenet21k} dataset using vision transformer~\cite{ViT}, and 4) DinoV2~\cite{dinov2}, pretrained on the LVD-142M dataset. For an equitable comparison, we standardized the use of ViT-Large with a 14-patch size and an input resolution of 336 pixels. As illustrated in Figure~\ref{fig:clip_siglippx} (left), it becomes evident that training on extensive visual corpora—akin to CLIP’s methodology—confers an advantage in terms of enhanced visual understanding capabilities.

\textit{Takeaway.} The overarching conclusion from our experiments in this section resonates with insights gained from the training of large vision-language models. In particular, we find that the vision backbone, trained using vision-language contrastive loss (i.e., CLIP and SigLIP), outperforms other models. Interestingly, our experiments further reveal that SigLIP demonstrates a surprisingly significant performance enhancement over CLIP, despite being supervised and operating on a considerably smaller training dataset. This observation suggests that refining training methodologies for CLIP-based models could be more impactful for small Vision-Language Models (VLMs) than for their larger counterparts, indicating a potential shift in optimization strategies for enhancing model effectiveness across different scales.
\\
\\
\noindent
\textbf{Scaling Image Resolution.}  Image resolution plays a vital role among various design choices in neural network architecture, with past research~\cite{scalingmllm} indicating that, for MLLMs, increasing image resolution is more effective than scaling other dimensions, such as model width and depth. This assertion is supported by findings~\cite{karamcheti2024prismatic} that enhancing image resolution uniformly improves model performance across all benchmarks. Naturally, one might assume that this principle should apply equally to MSLMs. However, is this assumption truly valid?

We present experiments using a popular visual backbone, CLIP, with varying image sizes. The experimental results are presented in Fig~\ref{fig:clip_siglippx} (middle).
Contrary to the common belief that increasing image resolution is always beneficial to model generalization, our empirical results indicate that higher image resolution does not guarantee improved performance for MSLMs. 
For instance, a CLIP model with a 224px resolution outperformed its higher-resolution counterparts in specific benchmarks, while a 336px resolution CLIP model excelled in others.

Despite these varied results, it's acknowledged that increased resolution can benefit tasks requiring detailed visual understanding, such as OCR, where higher resolution inputs enable more accurate recognition of details like price tags in our qualitative example in Fig~\ref{fig:clip_siglippx} (right). 
\\
\\
\noindent
\textit{Takeaway.} \textbf{Scaling up image resolution is not a silver bullet.} Although higher image resolutions offer more detailed visual inputs, they do not invariably enhance the performance of MSLMs. It is essential to find the right equilibrium between resolution and computational efficiency, tailored to the model's application context and the need for consistent understanding and output. Efforts might be better allocated to scaling other aspects of the visual backbone, such as its depth, rather than relying solely on increased image resolution.

\subsection{Optimization}
\begin{figure}[tb]
  \centering
  \includegraphics[width=0.7\linewidth]{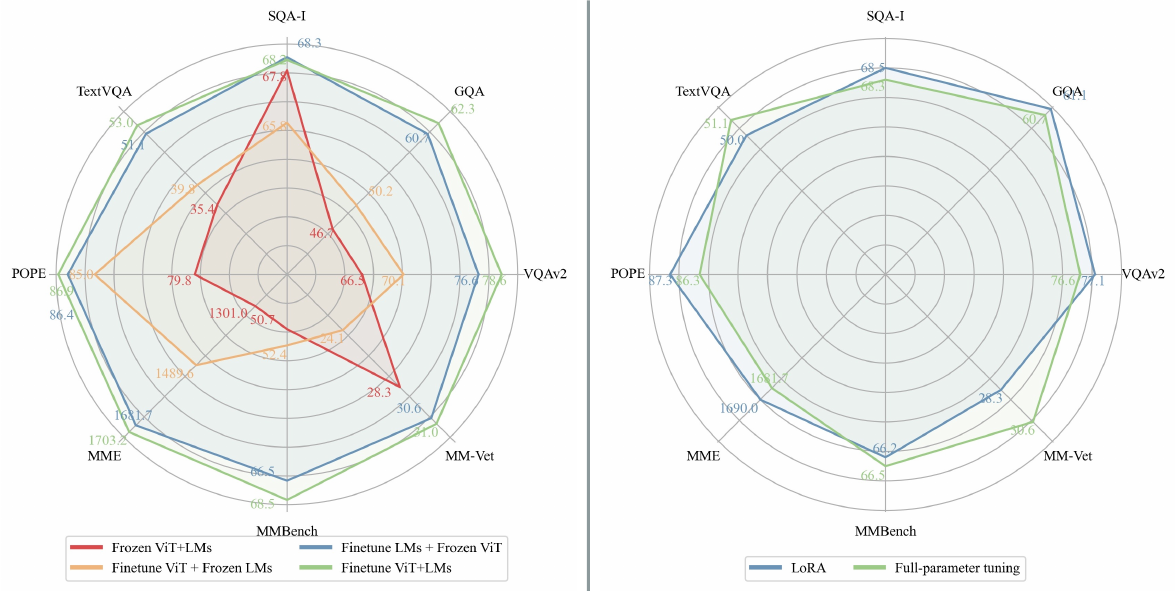}
  \caption{\textbf{Frozen vs. Finetuning \& Full-parameter tuning vs. LoRA.} We study the impact of optimization strategies on the performance of small multi-modal models. Specifically, we explored the effects of activating or freezing the visual representation backbone and LM during the instruction tuning phase on model performance (\textbf{left}). Additionally, we confirmed that compared to full-parameter tuning, applying LoRA to MSLMs is equally effective and can significantly alleviate the training burden (\textbf{right}).}
  \label{fig:optimization}
\end{figure}
Optimization strategies play a pivotal role in shaping the generalizability of a model. Most MLLMs~\cite{llava,llava1.5,minigpt4,mplugowl2} typically adopt a two-stage optimization pipeline. The first stage, or the pretraining stage, is dedicated to harmonizing vision and language features, establishing a solid foundation for multimodal understanding. 

\input{appendix/main_exp}

The subsequent stage focuses on instructing the model to follow users' instructions. This section delves into the training paradigms of MLLMs, exploring the necessity of finetuning the language model and visual backbones. 
Additionally, we evaluate the feasibility and effectiveness of parameter-efficient fine-tuning methods, such as LoRA, in this context.
\\
\\
\noindent
\textbf{Frozen vs. Finetuning.} This section examines whether the visual representation backbone and the language model should be configured as trainable or frozen. We present and analyze four different configurations that were explored.
\begin{itemize}
    \item Frozen ViT+SLM: In this configuration, the parameters of both the ViT and SLM are held constant, allowing the training to focus exclusively on aligning these two components via the projector.
    \item Frozen SLM + Finetune ViT: Here, we freeze the small language model's parameters and solely fine-tune the Vision Transformer.
    \item Finetune LM + Frozen ViT: This strategy involves freezing the ViT while training both the projector and the small language model, an approach exemplified by LLaVA-1.5~\cite{llava1.5}.
    \item Finetune ViT + SLM: This comprehensive approach activates all parameters within the MSLMs, permitting the model to simultaneously refine its visual processing capabilities and improve instruction-following proficiencies.
\end{itemize}
Our exploration into these strategies involves selectively activating or freezing the visual and language model during the instruction-tuning stage. As depicted in Fig~\ref{fig:optimization} (left), our findings suggest that freezing the language model can detrimentally affect the MSLMs' performance across various benchmarks. Furthermore, we observe that fully engaging all parameters of the MSLMs can lead to improvements on certain benchmarks.
\\
\\
\noindent 
\textit{Takeaway.} Contrary to previous research that suggested fine-tuning the vision backbone could lead to a significant decrease in performance across all benchmarks~\cite{karamcheti2024prismatic}, our study indicates that fine-tuning the vision encoder can actually enhance the performance of MSLMs across all evaluated benchmarks. We believe the underlying reason for this discrepancy is that SLMs possess fewer active neurons than LLMs. As a result, the visual representations in MSLMs must be more expressive than those in MLLMs to enable the language model to effectively comprehend image contents. Furthermore, our observations reveal that fine-tuning the language model is even more critical for achieving a successful MSLM implementation. We attribute this necessity to the fact that SLMs have less pre-trained data and fewer parameters, making the learning process to understand images with the assistance of visual backbones a pivotal challenge.
\\
\\
\noindent
\textbf{Full-Parameter Tuning vs. LoRA.} The aforementioned section shows that fine-tuning visual backbone and LMs are necessary for MSLMs. However, full-parameter fine-tuning entails a significant increase in training costs. We investigate the potential of parameter-efficient fine-tuning techniques as alternatives to conventional full-parameter tuning. In Fig~\ref{fig:optimization} (middle), we juxtapose the performance of models utilizing Low-Rank Adaptation (LoRA) against those subjected to full-parameter tuning. For the LoRA setup, we configure $r$ to be 128 and $\alpha$ to be 256. The experimental outcomes suggest that LoRA attains performance on par with its full-parameter counterparts across all evaluated benchmarks, positioning it as a viable and efficient alternative to full-parameter fine-tuning.

%% file: slm_table.tex
\begin{table*}[t]
  \centering
  \caption{Comparison of small language models on Massive Multitask Language Understanding (MMLU)~\cite{mmlu}, BIG-Bench Hard (BBH)~\cite{bbh}, GSM8K~\cite{GSM8K}, and HumanEval~\cite{Humaneval}.}
  \label{tbl:slm_table}
  \resizebox{0.9\linewidth}{!}{
      \begin{tabular}{l|c|c|c|c}
        \toprule
         \multirow{2}{*}{Method} & MMLU  & BBH & GSM8K & HumanEval\\
         & Language Understanding & Instruction Following & Math & Coding \\
        \midrule
        Pythia-1.4B & 25.4 & - & - & 4.3\\
        StableLM-1.6B & 45.9 & 37.7 & 52.5 & 35.4\\
        Qwen1.5-1.8B & 46.8 & 24.2 & 38.4 & 20.1  \\
        Phi-1.5-1.3B & 43.9 & - & 44.6 & 41.4 \\
        \midrule
        Pythia-2.8B & 28.3 & - & - & 5.1  \\
        Phi-2-2.7B & 56.7 & 43.4 & 57.2 & 47.6  \\
        \bottomrule
      \end{tabular}
    }
\end{table*}

%% file: appendix/main_exp.tex
\begin{table*}[tb]
  \centering
  \caption{Multi-modal evaluation on 9 benchmarks: $\text{VQA}{\text{v2}}$~\cite{vqav2}; GQA~\cite{gqa}; $\text{SQA}^{\text{I}}$: ScienceQA-IMG~\cite{scienceqa}; $\text{VQA}^{\text{T}}$: TextVQA~\cite{textvqa}; \text{MME-P}: MME perception~\cite{mme}; \text{MME-C}: MME cognition~\cite{mme}; \text{MMB}: MMBench~\cite{mmbench}; \text{MM-Vet}~\cite{mmvet}; \text{POPE}~\cite{pope}. {$^*$}The training images/annotations of the datasets are observed during training. {$^ \dagger$} Includes in-house data that is not publicly accessible. Column \textit{Res.} is the image resolution of visual backbone. Columns \textit{PT} and \textit{IT} are the data sizes in the pretraining stage and the visual instruction tuning stage, respectively.}
  \label{tbl:full_table}
  \resizebox{1.0\linewidth}{!}{
      \begin{tabular}{l|l|l|l|l|ccccccccc}
        \toprule
           Method & LM  & Res. & PT & IT & $\text{VQAv2}$ &\text{GQA} & \text{SQA$^{\text{I}}$} & $\text{VQA}^{\text{T}}$ & \text{MME-P} & \text{MME-C} & \text{MMB} & \text{MM-Vet} & \text{POPE} \\
        \midrule
        \multicolumn{14}{c}{Multimodal Large Language Models} \\
        \midrule
        BLIP-2~\cite{blip-2} & Vicuna (13B) & 224 & 129M & -                               & 65.0 & 41.0 & 61.0 & 42.5 & 1293.8 & 290.0 & - & 22.4 & 85.3\\
        InstructBLIP~\cite{instructblip} & Vicuna (7B) & 224 & 129M & 1.2M                 & - & 49.2 & 60.5 & 50.1 & - & - & 36 & 26.2 & - \\
        InstructBLIP~\cite{instructblip} & Vicuna (13B) & 224 & 129M & 1.2M                & - & 49.5 & 63.1 & 50.7 & 1212.8 & 291.8 & - & 25.6 & 78.9 \\
        Shikra~\cite{shikra} & Vicuna (13B) & 224 & 600K & 5.5M                            &  77.4{$^*$} & - & - & - & - & - & 58.8 & - & -\\
        IDEFICS-9B~\cite{idefics} & LLaMA (7B) & 224 & 353M & 1M                           & 50.9 & 38.4 & - & 25.9 & - & - & 48.2 & - & -  \\
        IDEFICS-80B~\cite{idefics} & LLaMA (65B) & 224 & 353M & 1M                         & 60.0 & 45.2 & - & 30.9 & - & - & 54.5 & - & - \\
        Qwen-VL~\cite{qwen} & Qwen (7B) & 448 & 1.4B{$^\dagger$} & 50M{$^\dagger$}         & 78.8{$^*$} & 59.3{$^*$} & 67.1 & \textbf{63.8}{$^*$} & - & - & 38.2 & - & -\\
        Qwen-VL-Chat~\cite{qwen} & Qwen (7B) & 448 & 1.4B{$^\dagger$} & 50M{$^\dagger$}    & 78.2{$^*$} & 57.5{$^*$} & 68.2 & 61.5{$^*$} & 1487.5 & \textbf{360.71} & 60.6 & - & -\\
        mPLUG-Owl2~\cite{mplugowl2} & LLaMA (7B) & 448 & 400M & 1.23M                      & 79.4 & 56.1 & 68.7 & 58.2 & 1450.2 & 313.2 & 64.5 & \textbf{36.2} & 85.8 \\
        LLaVA-1.5~\cite{llava1.5} & Vicuna (7B) & 336 & 558K & 665K                        & 78.5{$^*$} & 62.0{$^*$} & 66.8 & 58.2 & 1510.7 & 316.1 & 64.3 & 30.5 & 85.9 \\
        LLaVA-1.5~\cite{llava1.5} & Vicuna (13B) & 336 & 558K & 665K                       & 80.0{$^*$} & 63.3{$^*$} & \textbf{71.6} & 61.3 & \textbf{1531.3} & 295.4 & 67.7 & 35.4 & 85.9 \\
        \midrule
        \multicolumn{14}{c}{Multimodal Small Language Models} \\
        \midrule
        MobileVLM-1.7B~\cite{chu2023mobilevlm} & M-LLaMA (1.4B) & 336 & 558K & 665K        & - & 56.1 &  57.3 & 41.5 & 1196.2 & - & 53.2 & - & 84.5 \\ 
        MobileVLM-3B~\cite{chu2023mobilevlm} & M-LLaMA (2.7B) & 336 & 558K & 665K          & - & 59.0 & 61.2 & 47.5 & 1288.9 & - & 59.6 & - & 84.9 \\
        MobileVLM-v2-1.7B~\cite{mobilevlmv2} & M-LLaMA (1.4B) & 336 & 1.2M & 2.4M          & -  & 59.3 & 66.7 & 52.1 & 1302.8 & - & 57.7 & - & 84.3\\
        MobileVLM-v2-3B~\cite{mobilevlmv2} & M-LLaMA (2.7B) & 336 & 1.2M & 2.4M            & -  & 61.1 & 70.0 & 57.5 & 1440.5 & - & 63.2 & - & 84.7 \\
        MoE-LLaVA-3.6B~\cite{moe-llava} & Phi-2 (2.7B) & 384 & 558k & 1.59M                & 79.9{$^*$} & 62.6{$^*$} & 70.3 & 57.0 & 1431.3 & - & 68.0 & 35.9 & 85.7 \\
        Bunny-3B~\cite{bunny} & Phi-2 (2.7B) & 384 & 2M & 695K                             & 79.8 & 62.5 & 70.9 & - & 1488.8 & 289.3 & 68.6 & - & 86.8\\
        \midrule
        Mipha-1.6B & Phi-1.5 (1.3B) & 384 & 558K & 665K                                    & 77.5{$^*$} & 62.7{$^*$} & 58.3 & 45.6 & 1203.1 & 247.9	& 57.7 & 23.5 & \textbf{86.9} \\
        Mipha-2.4B & Gemma (2.0B) & 384 & 558K & 665K                                      & 79.5{$^*$} & 63.3{$^*$} & 65.3 & 52.4 & 1397.1 & 265.7 & 59.4 & 29.9 & 86.6 \\
        Mipha-3B & Phi-2 (2.7B)  & 384 & 558K & 665K                                       & \textbf{81.3}{$^*$} & \textbf{63.9}{$^*$} & 70.9 & 56.6 & 1488.9 & 295.0 & \textbf{69.7} & 32.1 & 86.7\\
        \bottomrule
      \end{tabular}
    }
\end{table*}

%% file: experiment.tex
\section{Experiments}
\subsection{Implementation Details}
\label{sec:kds}
\textbf{Integration of Key Design Elements.} Through our empirical study, we have distilled a set of insights that collectively enhance both the downstream performance and the efficiency of models:
\begin{itemize}
    \item \textit{Visual Representations.} We have chosen the SigLIP backbone with a resolution of 384px as our standard for visual processing.
    \item \textit{Language Models.} We use Supervised finetuned Phi-2-2.7B as our language backbone for Mipha-3B.
    \item \textit{Optimization.} Recognizing the importance of fine-tuning both the visual and language components, we undertake fine-tuning of both models using their full parameter sets during the training phase
\end{itemize}
Leveraging these insights, we developed a novel suite of models, named Mipha, scaled to 3 and 1.7 billion parameters, aiming to set new benchmarks in multimodal models. More detailed training settings are presented in the appendix.

\subsection{Comparison with SOTA Models}
We evaluate the visual question answering abilities on VQAv2~\cite{vqav2}, GQA~\cite{gqa}, SQA{$^{\text{I}}$}~\cite{scienceqa} and VQA{$^{\text{T}}$}~\cite{textvqa}. We also tested against more challenging benchmarks designed for MLLMs, i.e., POPE~\cite{pope}, MME~\cite{mme}, MMBench~\cite{mmbench}, and MM-Vet~\cite{mmvet}. As shown in Table~\ref{tbl:full_table}, Mipha-3B outperformed other models in three of the nine benchmarks, despite being compared to models with twice its size. Specifically, Mipha-3B surpassed similar-sized multimodal models like Bunny-3B and MobileVLM-v2-3B in six benchmarks, despite having only 45\% of the training data of Bunny-3B and 34\% of MobileVLM-v2-3B. Moreover, our Mipha-2.4B and Mipha-1.6B models demonstrated competitive performance across these benchmarks. For example, Mipha-2.4B exceeded the performance of other smaller multimodal models, such as MoE-LLaVA-3.6B, particularly in the GQA benchmark.


%% file: conclusion.tex
\section{Conclusion}
In this paper, we delve deeply into three aspects of Multimodal Small Language Models (MSLMs): the language models, visual representations, and optimization strategy. On the basis of our thorough analysis, we identify several improvements that can enhance the performance of MSLMs. Integrating insights from these aspects, we have developed a novel suite of models called Mipha. Compared to other MSLMs, our top-performing model, Mipha-3B, secures the highest performance in 8 out of 11 benchmarks. Moreover, when compared to the current state-of-the-art model, LLaVA-1.5-13B, Mipha-3B demonstrates comparable performance. We believe that Mipha offers a novel perspective on the training of strong MSLMs.

%% file: appendix/appendix.tex
\appendix
\section{Limitation}
While this work has extensively and thoroughly explored the design space of Multimodal Small Language Models (MSLMs), our analysis predominantly draws upon methodologies from current models, such as Phi-1.5/2~\cite{phi-1.5, Phi-2}, StableLM~\cite{stablelm}, Pythia~\cite{pythia}, and Gemma~\cite{gemma}, which are considered smaller language models. Given the rapid evolution of the field, the findings and conclusions of our study might not remain applicable to newer language models that emerge. Moreover, we recognize significant potential for further enhancements in Mipha, particularly in areas like Optical Character Recognition (OCR) and mathematical problem-solving. We plan to tackle these shortcomings in future research endeavors.

\section{Negative Social Impact}
Despite Mipha showcasing advanced multimodal comprehension capabilities, it is not immune to the negative societal impacts observed in existing multimodal models. Foremost, Mipha has the potential to produce unreliable responses and may inadvertently perpetuate societal biases, including gender biases. Additionally, there exists the risk of Mipha being manipulated by users to generate harmful content. 

\section{Implementation Details}
The training process and dataset adopted by Mipha are substantially identical to LLaVA-1.5~\cite{llava1.5}. Specifically, Mipha adopts a two-stage training pipeline, i.e., the pretraining stage and the instruction tuning stage. Initially, we align the vision features from SigLIP~\cite{siglip} with the text embedding from Small Language Models (SLMs)~\cite{phi-1.5,Phi-2}. Subsequently, we apply visual instruction tuning to fully harness the capabilities of the Multimodal Small Language Models (MSLMs) across various multimodal tasks. We adopt the same cross-entropy loss during both stages for the next token prediction. In the pretraining stage, only the parameters of projector (a two-layer MLP) are updated on LCS-558K~\cite{llava1.5}. The learning rate of 1e-3 and a batch size of 256 are adopted at this stage. In the instruction tuning stage, we train all parameters of Mipha for 2 epochs on a collection of mixture datasets with 665K samples, which is organized by~\cite{llava1.5}. At this stage, we set the learning rate as 2e-5 and the batch size as 128. For both stages, we utilize the AdamW optimizer with no weight decay, characterized by momentum parameters of 0.9 and 0.98, and an epsilon value of 1e-7. We provide overall training settings in Fig~\ref{tab:hyperparam}.
Additionally, for our model variants, we employ the same hyperparameters as Mipha, except for setting the epoch to 1 during the finetuning stage. We employ publicly available datasets to conduct instruct tuning for LMs. Specifically, we utilize the Alpaca dataset~\cite{alpaca} to supervised fine-tune Phi-2. For RHLF~\cite{rlhf} and Safe-RLHF~\cite{safe-rlhf}, we employ the PKU-SafeRLHF-10K dataset~\cite{safe-rlhf} to finetune the models.
\input{appendix/training_setting}
\begin{figure}[t]
  \centering
  \includegraphics[width=0.75\linewidth]{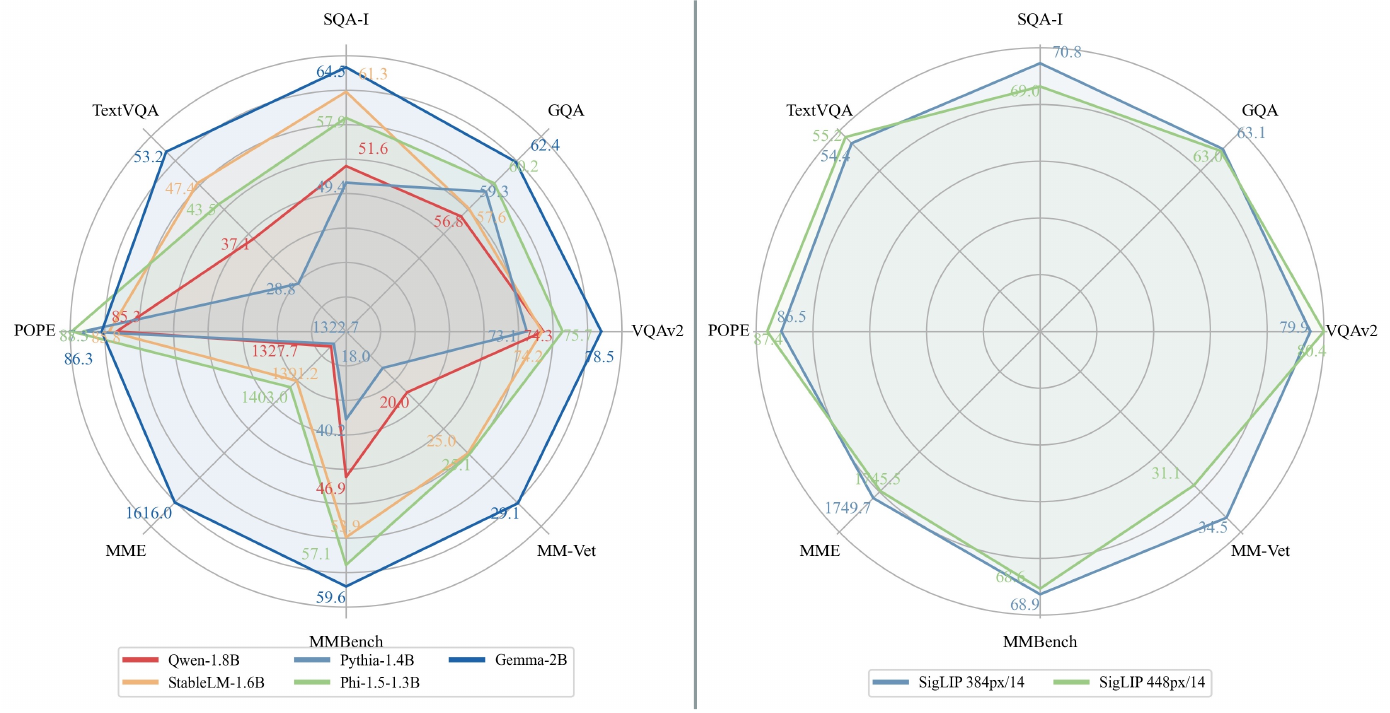}
  \caption{\textbf{Additional Experimental Results: Selection of Small Language Models (Left) \& Scaling Image Resolution (Right).}}
  \label{fig:appendix_selection_slm}
\end{figure}

\noindent

\subsection{More Experiments}

\subsubsection{Selection of Small Language Models.}
Beyond the experiments presented in Fig. ~\ref{fig:slm} (left) of the main text, we have expanded our evaluation to include the performance of the Gemma-2B model~\cite{gemma}. We compared Gemma-2B with several smaller language models: Qwen-1.8B~\cite{bai2023qwen}, Pythia-1.4B~\cite{pythia}, StableLM-1.6B~\cite{stablelm}, and Phi-1.5-1.3B~\cite{phi-1.5}. The results of these comparisons are illustrated in Fig~\ref{fig:appendix_selection_slm} (left). The findings suggest that multimodal models leveraging Gemma-2B are versatile, achieving top-tier performance across the majority of benchmarks when compared to alternative methods. As one of the most recent additions to the small language model category, its performance on language-based tasks has exceeded that of LLaMA-2~\cite{llama2}. We posit that Gemma-2B's robust performance makes it an excellent foundational model for the development of MSLMs.

\subsubsection{Scaling Image Resolution.}
In addition to CLIP~\cite{clip}, we also conduct experiments using SigLIP~\cite{siglip} with resolutions of 384px and 448px. Our findings, as illustrated inFig~\ref{fig:appendix_selection_slm} (right), reveal that the SigLIP model with a 384px resolution outperforms its 448px counterpart in five out of eight benchmarks~\cite{gqa,textvqa,vqav2,scienceqa,pope,mmvet,mmbench,mme}. This suggests that simply increasing image resolution does not consistently improve the model’s performance across a diverse set of multimodal tasks. Additionally, with the exception of MM-Vet~\cite{mmvet}, the two SigLIP variants show minimal variance in performance on the other seven benchmarks. This observation reinforces our assertion that higher image resolutions do not necessarily equate to better results in MSLMs.

\subsubsection{More Experimental Result on Mipha.}
\input{instruct_following_table}

\textbf{Experimental Results on Visual Question Answering Benchmarks.} We evaluate the visual question answering abilities on VQAv2~\cite{vqav2}, GQA~\cite{gqa}, VizWiz~\cite{vizwiz}, SQA{$^{\text{I}}$}~\cite{scienceqa} and VQA{$^{\text{T}}$}~\cite{textvqa}. As shown in Table~\ref{tbl:main}, Mipha-3B achieves the highest performance in 2 out of the 5 benchmarks.
Notably, when compared to Bunny-3B~\cite{bunny}, which is trained on four times more data than Mipha-3B, our method demonstrates superior performance with improvements of 1.5\% and 1.4\% on VQAv2 and GQA, respectively. This underlines the effectiveness of our optimization strategies. Moreover, without the need for an expanded training dataset, Mipha-3B stands out on VQAv2 and GQA, outperforming LLaVA-1.5-13B~\cite{llava1.5} on VQAv2 by 1.3\% and on GQA by 0.6\%. 
These findings underscore that Mipha-3B, despite having only 23\% of the parameter count of LLaVA-1.5-13B, can still achieve commendable generalization in certain scenarios.
\\
\\
\noindent
\textbf{Experimental Results on Instruction-Following Benchmarks.} Furthermore, we evaluate our proposed Mipha on more challenging benchmarks that are designed for evaluation of MLLMs, i.e., POPE~\cite{pope}, MME~\cite{mme}, MMBench~\cite{mmbench}. MMBench-CN~\cite{mmbench}, SEED-Bench-img~\cite{seed}, and MM-Vet~\cite{mmvet}. As shown in Table~\ref{tbl:qa_table}, our proposed model, Mipha-3B, exceeds LLaVA-1.5-13B in performance on MMBench and MMBench-CN by 2\% and 0.7\%, respectively, despite both models being trained on the same dataset. 
Additionally, Mipha-3B demonstrates stronger resistance to hallucinations, surpassing LLaVA-1.5-13B on POPE by 0.8\%. 
In comparison with other MSLMs, Bunny-3B, MoE-LLaVA-3.6B~\cite{moe-llava}, and MobileVLM-v2-3B~\cite{mobilevlmv2}, we are astonished to find that despite Mipha-3B utilizing the least amount of training data, it achieved the best performance in 4 out of 6 benchmarks. 
In particular, on MMBench, Mipha-3B outperformed Bunny-3B by 1.1\% and MoE-LLaVA by 1.7\%, which are combined with the same visual backbone SigLIP and LM Phi-2-2.7B respectively.  These results validate the effectiveness of our optimization strategies.



\subsection{Qualitative Examples}
\begin{figure}[h]
  \centering
  \includegraphics[width=\linewidth]{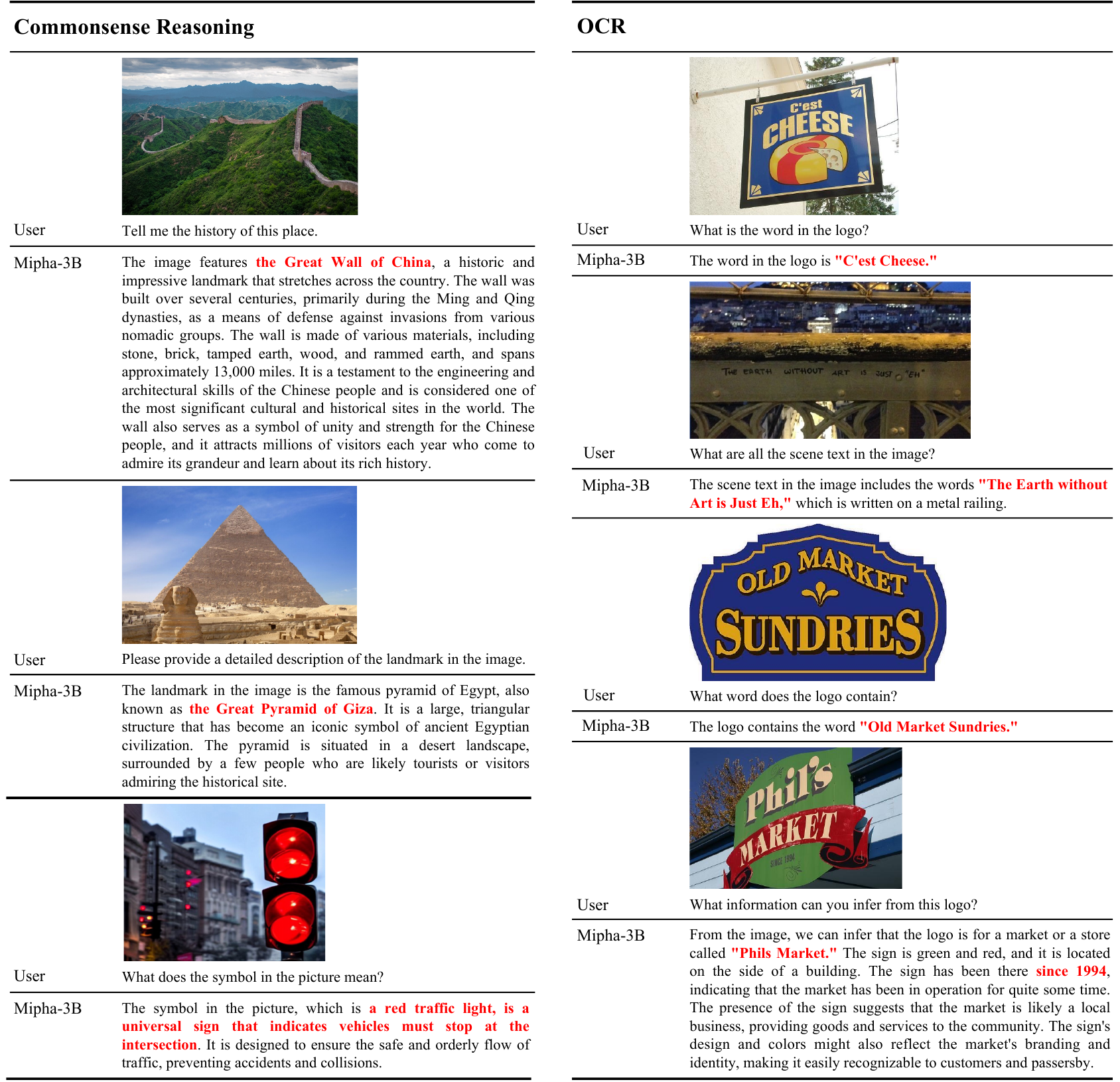}
  \caption{\textbf{Some cases of Commonsense Reasoning \& OCR by Mipha-3B.} For Commonsense Reasoning, Mipha-3B successfully identifies the Great Wall of Chain and the Great Pyramid of Giza in the image, going further to provide more detailed descriptions of them. Additionally, Mipha-3B also demonstrates strong OCR capabilities, such as successfully recognizing ``since 1994'' in the bottom right subfigure. \textcolor{red}{\textbf{RED}} means the critical part of the answer.}
  \label{fig:supp_examples_1}
\end{figure}

\begin{figure}[h]
  \centering
  \includegraphics[width=0.95\linewidth]{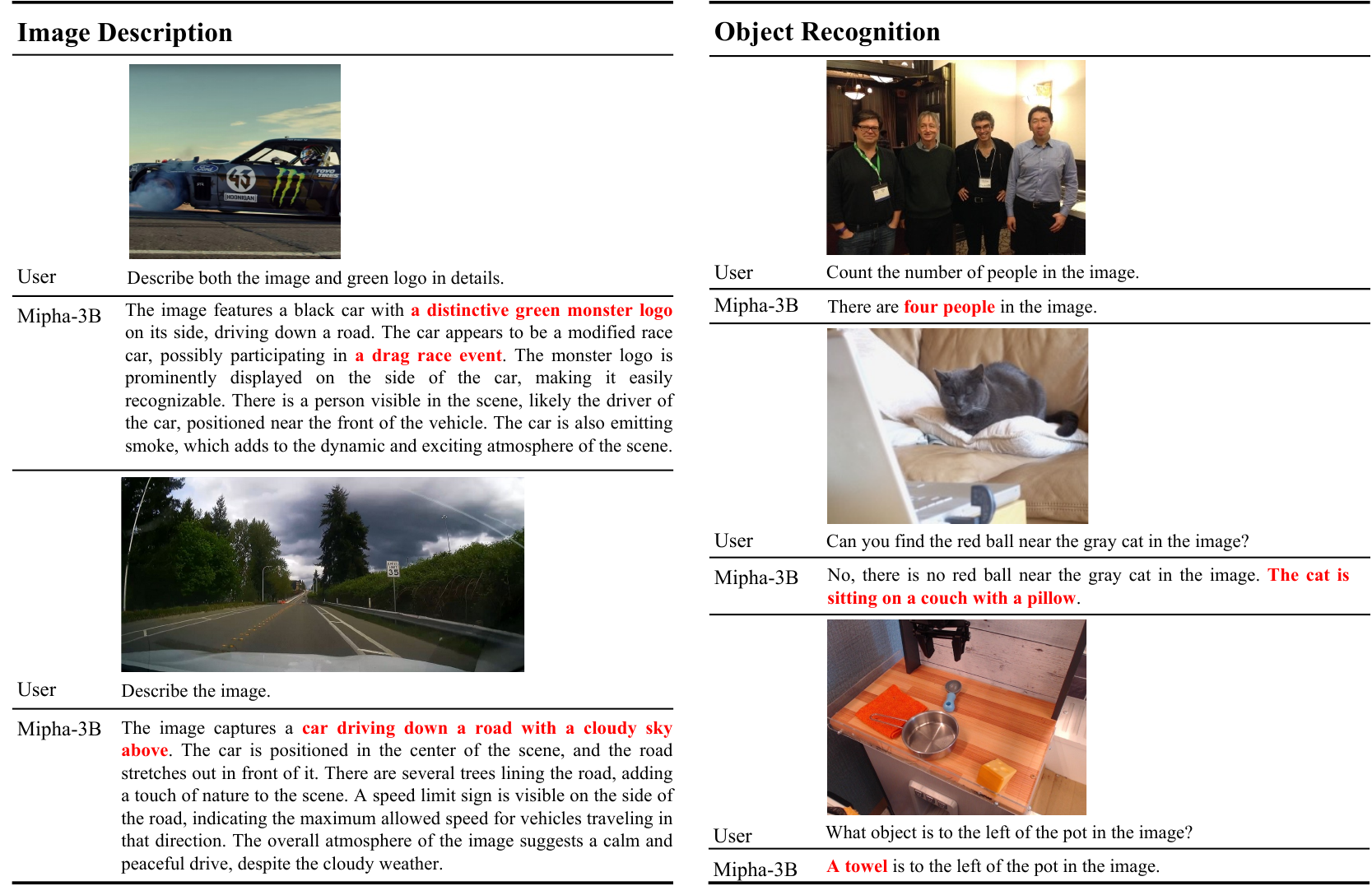}
  \caption{\textbf{Some cases of Image Description \& Object Recognition by Mipha-3B.} Mipha-3B is capable of inferring potential occurrences based on the scene, such as ``a drag race event'', showcasing its robust visual perception abilities. Additionally, Mipha-3B demonstrates strong object recognition capabilities, including aspects such as counting, existence, and orientation. \textcolor{red}{\textbf{RED}} means the critical part of the answer.}
  \label{fig:supp_examples_2}
\end{figure}

\begin{figure}[h]
  \centering
  \includegraphics[width=0.95\linewidth]{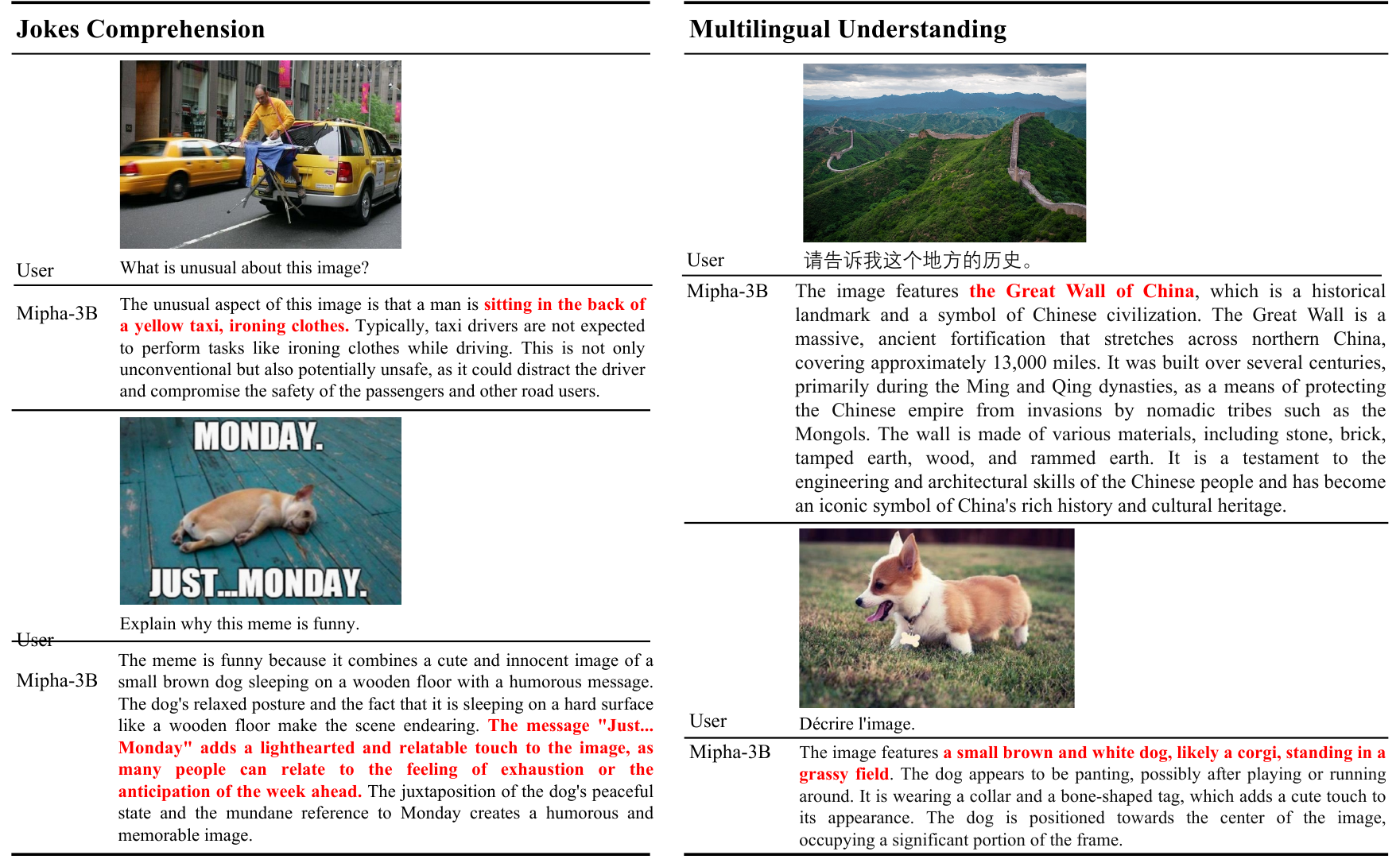}
  \caption{\textbf{Some cases of Jokes Comprehension \& Multilingual Understanding by Mipha-3B.} Mipha-3B comprehends intriguing elements within the image and demonstrates multilingual abilities across Chinese and French. \textcolor{red}{\textbf{RED}} means the critical part of the answer.}
  \label{fig:supp_examples_3}
\end{figure}

In Fig~\ref{fig:supp_examples_1}, \ref{fig:supp_examples_2} and \ref{fig:supp_examples_3}, we present several cases from various perspectives such as commonsense reasoning, OCR, image description, object recognition, joke comprehension, and multilingual understanding. Mipha-3B demonstrates powerful detailed visual understanding by comprehensively answering questions based on the information in the images.

%% file: appendix/training_setting.tex
\begin{table*}[t]
    \centering
    \caption{Hyper-parameters for training Mipha.}
    \resizebox{0.7\linewidth}{!}{
    \begin{tabular}{l|cc}
         \toprule
         Configuration            & Pretraining & Instruction Tuning \\
         \midrule
         ViT init.                & SigLIP 384px/14 & SigLIP 384px/14 \\
         Projector init.          & Random & Pretrain stage \\
         Image resolution         & \multicolumn{2}{c}{$384\times 384$} \\
         ViT sequence length      & \multicolumn{2}{c}{729} \\
         LLM sequence length      & \multicolumn{2}{c}{2048} \\
         Optimizer                & \multicolumn{2}{c}{AdamW} \\
         Optimizer hyperparameter & \multicolumn{2}{c}{$\beta_{1}=0.9, \beta_{2}=0.98, \epsilon=1e^{-7}$} \\
         Peak learning rate       & $1e^{-4}$ & $2e^{-5}$ \\
         Training Epochs          & 1 & 2 \\
         Global batch size        & 256 & 128 \\
         Learning rate schedule   & \multicolumn{2}{c}{Cosine} \\
         Weight decay             & \multicolumn{2}{c}{0} \\
         Warm-up ratio            & \multicolumn{2}{c}{0.03} \\
         Numerical precision      & \multicolumn{2}{c}{$\mathtt{bfloat16}$} \\
         Activation checkpointing & \multicolumn{2}{c}{\checkmark} \\
         \bottomrule
    \end{tabular}
    \label{tab:hyperparam}}
\end{table*}

%% file: instruct_following_table.tex
\begin{table*}[tb]
  \centering
  \caption{Multi-modal evaluation on instruction following benchmarks. POPE~\cite{pope}; MME-P: MME Perception~\cite{mme}; MME-C: MME Cognition~\cite{mme}; MMB: MMBench~\cite{mmbench}; SEED: SEED-Bench~\cite{seed}; MM-Vet~\cite{mmvet}. {$^ \dagger$} Includes in-house data that is not publicly accessible. Columns \textit{PT} and \textit{IT} are the data sizes in the pretraining stage and the visual instruction tuning stage, respectively.}
  \label{tbl:main}
  \resizebox{1.0\linewidth}{!}{
      \begin{tabular}{l|l|l|c|cc|cc|c|c}
        \toprule
        \multirow{2}{*}{Method} & \multirow{2}{*}{PT} & \multirow{2}{*}{IT} & \multirow{2}{*}{POPE} & \multicolumn{2}{|c|}{MME-} & \multicolumn{2}{|c|}{MMB} & \multicolumn{1}{|c|}{SEED} & \multirow{2}{*}{MM-Vet} \\
         & & & & P & C & en & cn & img & \\
        \midrule
        \multicolumn{10}{c}{Multimodal Large Language Models} \\
        \midrule
        BLIP2-14B~\cite{blip-2} & 129M & - & 85.3 & 1293.8 & 290.0 & - & - & 49.7 & 22.4\\
        InstructBLIP-8B~\cite{instructblip} & 129M & 1.2M & - & - & - & 36 & 23.7 & 58.8 & 26.2 \\
        InstructBLIP-14B~\cite{instructblip} & 129M & 1.2M& 78.9 & 1212.8 & 291.8 & - & - & - & 25.6 \\
        Shikra-13B~\cite{shikra} & 600K & 5.5M & - & - & - & 58.8 & - & - & - \\
        IDEFICS-80B~\cite{idefics} & 353M & 1M & - & - & - & 54.5 & 38.1 & 53.2 & - \\
        Qwen-VL~\cite{qwen} & 1.4B$^{\dagger}$ & 50M$^{\dagger}$ & - & - & - & 38.2 & 7.4 & 62.3 & - \\
        Qwen-VL-Chat~\cite{qwen} & 1.4B$^{\dagger}$ & 50M$^{\dagger}$ & - & 1487.5 & \textbf{360.71} & 60.6 & 56.7 & 65.4 & - \\
        mPLUG-Owl2-7B~\cite{mplugowl2} & 400M & 1.23M & 85.8 & 1450.2 & 313.2 & 64.5 & - & - & \textbf{36.2}\\
        LLaVA-7B~\cite{llava}  & 595K & 158K & 72.9 & 809.6 & 247.9 & 38.7 & 36.4 & 37.0 & 25.5 \\
        LLaVA-1.5-7B~\cite{llava1.5} & 558K & 665K & 85.9 & 1510.7 & 316.1 & 64.3 & 58.3 & 66.1 & 30.5 \\
        LLaVA-1.5-13B~\cite{llava1.5}& 558K & 665K & 85.9 & \textbf{1531.3} & 295.4 & 67.7 & 63.6 & 68.2 & 35.4\\
        \midrule
        \multicolumn{10}{c}{Multimodal Small Language Models} \\
        \midrule
        MobileVLM-3B~\cite{chu2023mobilevlm} & 558K & 665K & 84.9 & 1288.9 & - & 59.6 & - & - & -\\
        MobileVLM-v2-3B~\cite{mobilevlmv2} & 1.2M & 2.4M & 84.7 & 1440.5 & -  & 63.2 & - & -  & -\\
        MoE-LLaVA-3.6B~\cite{moe-llava} & 558K & 1.59M & 85.7 & 1431.3 & - & 68.0 & - & - & 35.9 \\
        Bunny-3B~\cite{bunny} & 2M & 695K & \textbf{86.8} & 1488.8 & 289.3 & 68.6 & - & - & - \\
        \midrule
        Mipha-3B (Ours) & 558K & 665K & 86.7 & 1488.9 & 295.0 & \textbf{69.7} & \textbf{64.3} & \textbf{68.9} & 32.1 \\
        \bottomrule
      \end{tabular}
    }
\end{table*}